%% file: cvpr.tex
\documentclass[10pt,twocolumn,letterpaper]{article}

\usepackage{cvpr}

\input{preamble}

\definecolor{cvprblue}{rgb}{0.21,0.49,0.74}

\usepackage{times}
\usepackage{epsfig}
\usepackage{graphicx}
\usepackage{amsmath}
\usepackage{amssymb}
\usepackage{enumitem}
\usepackage{multirow}
\usepackage{float}
\usepackage{makecell}
\usepackage[section]{placeins}
\usepackage{mathtools}
\usepackage[normalem]{ulem}
\usepackage{gensymb}

\usepackage[pagebackref=true,breaklinks=true,colorlinks,bookmarks=false]{hyperref}

\def\paperID{6} 
\def\confName{CVPR}
\def\confYear{2024}

\begin{document}

\title{Affine-based Deformable Attention and \\
 Selective Fusion for Semi-dense Matching}

\author{Hongkai Chen$^{1}$\hspace{0.15cm} Zixin Luo$^{1}$\hspace{0.15cm} Yurun Tian$^{1}$\hspace{0.15cm}Xuyang Bai$^{1}$\hspace{0.15cm}Ziyu Wang$^{1}$\hspace{0.15cm} Lei Zhou$^{1}$\hspace{0.15cm}Mingmin Zhen\hspace{0.15cm}  \\Tian Fang$^{1}$\hspace{0.15cm}David McKinnon$^{1}$\hspace{0.15cm} Yanghai Tsin$^{1}$\hspace{0.15cm} Long Quan$^{2}$ \\
\normalsize $^1$Apple Inc. \hspace{0.7cm} $^2$Hong Kong University of Science and Technology \hspace{0.7cm} \normalsize\\
\tt\small
}

\maketitle

\begin{abstract}
Identifying robust and accurate correspondences across images is a fundamental problem in computer vision that enables various downstream tasks. Recent semi-dense matching methods emphasize the effectiveness of fusing relevant cross-view information through Transformer. In this paper, we propose several improvements upon this paradigm. Firstly, we introduce affine-based local attention to model cross-view deformations. Secondly, we present selective fusion to merge local and global messages from cross attention. Apart from network structure, we also identify the importance of enforcing spatial smoothness in loss design, which has been omitted by previous works. Based on these augmentations, our network demonstrate strong matching capacity under different settings. The full version of our network achieves state-of-the-art performance among semi-dense matching methods at a similar cost to LoFTR, while the slim version reaches LoFTR baseline's performance with only 15\% computation cost and 18\% parameters.

\end{abstract}

\let\saveFloatBarrier\FloatBarrier
\let\FloatBarrier\relax

\section{Introduction}
\let\FloatBarrier\saveFloatBarrier
Robust and accurate image matching serves as a critical front-end task for a wide range of applications that require estimating geometry from RGB input, such as Structure-from-Motion (SfM)~\cite{schonberger2016structure,sfm2}, Simultaneous Localization And Mapping (SLAM)~\cite{mur2015orb,orbslam2} and Visual Localization~\cite{sattler2012imageBMCV}. Conventionally, image matching is comprised of several individual stages, including keypoint extraction, feature description, and feature matching. In the past few years, the research community has observed a remarkable progress in replacing traditional steps with their learning-based counterparts~\cite{geodesc,luo2020aslfeat,detone2018superpoint,oanet,acnet,d2net,lift}, leading to promising improvements. More recently, increasing efforts have been made towards more unified and end-to-end image matching systems, which are typically implemented in dense or semi-dense fashion by incorporating Transformer structure~\cite{sun2021loftr,chen2022aspanformer,tang2022quadtree,wang2022matchformer,topicfm}, cost-volume regularization~\cite{pdcnet+,ncnet,truong2021pdc,sparsencnet,li2020dual} and coarse-to-fine scheme~\cite{sun2021loftr,dkm,roma}, which process image pairs directly and bypass limitations imposed by pre-extracted keypoints, such as repeatability issue and inability to handle low-texture areas. 

Specifically, in Transformer-based matchers, striking a balance between token granularity and computation efficiency is crucial. To address this challenge, recent works~\cite{chen2022aspanformer,astr} propose global-local attention frameworks which utilizes global attention at a coarse level to model long-range dependencies, while local attention facilitates fine-level message exchange. Although these methods demonstrate the ability to concentrate attention span into specific areas, we consider below limitations still persist.

On the one hand, the local attention usually adopts rectangular grid areas to sample tokens, disregarding complex local deformation. Due to the nature of two-view matching tasks, corresponding regions in image pairs are usually related by some extent of deformations, including scaling, shearing and rotation. Consequently, a rectangular patch in the source view will be projected to a deformed area in the target image, which should be considered in token sampling for local attention to maximize the overlap ratio between sampling areas in source/target feature maps.

On the other hand, the global-local message fusion in previous works are usually conducted through learned priors~\cite{astr,chen2022aspanformer,tang2022quadtree}, without considering the differences in reliability among sampled patches. Due to the imprecision in intermediate flow estimation and the existence of non-overlapping areas, accepting all local message equally introduces noise for feature update, not to mention the local message from non-overlapped regions. In this regard, an ideal global-local message fusion should suppress local information from unreliable or non-related local areas.

\begin{figure*}
    \centering 
	\includegraphics[width=1\textwidth]{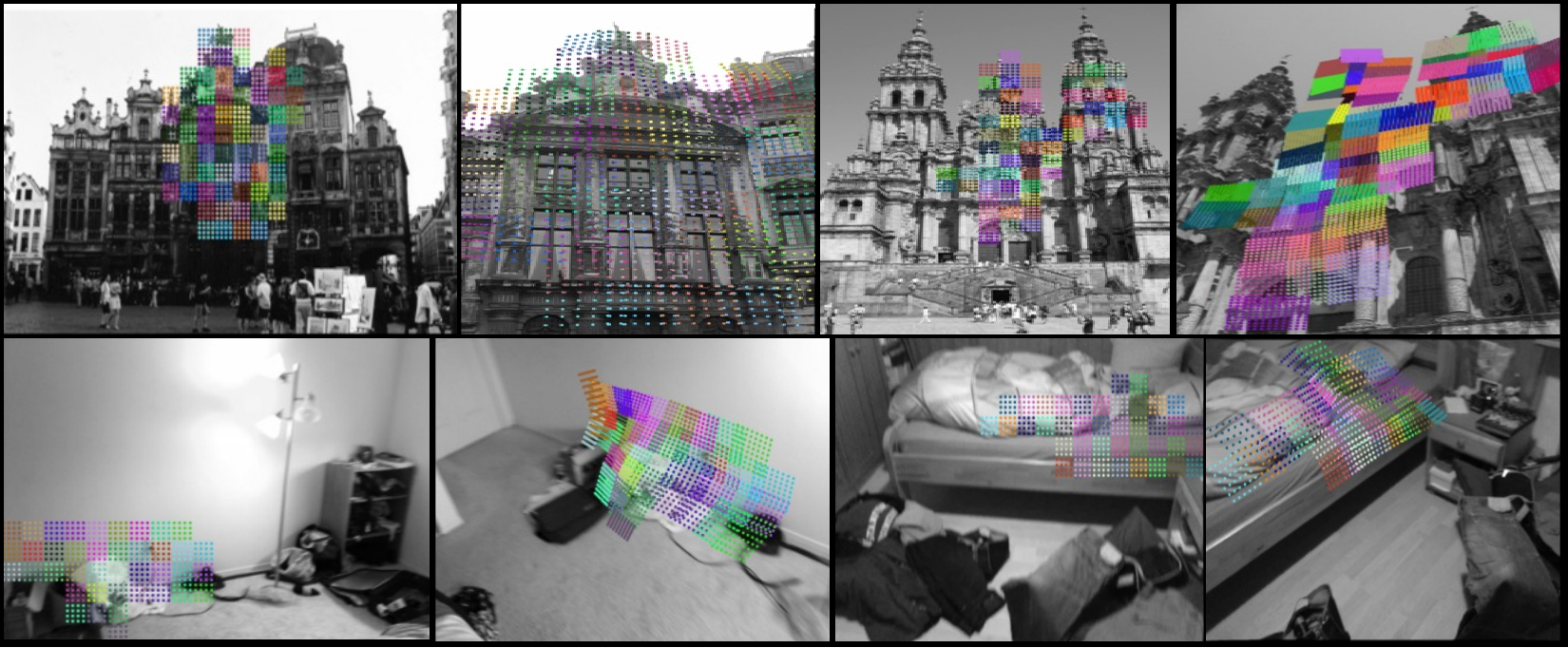}
	\caption{Visualization of the proposed deformable attention. Through piece-wise deformation estimation, we project source patches (left) to the target image (right) to sample tokens in local attention.}
	\label{deform}
\end{figure*}

Apart from network designs, we also review the de-factor loss design for semi-dense matching methods~\cite{sun2021loftr,eloftr,tang2022quadtree,chen2022aspanformer,wang2022matchformer}. In this series of works, classification-based loss (such as focal loss) is applied to assignment matrix to maximize (minimize) positive (negative) entries. Although classification-based loss is essential for learning distinctive features, it should be noticed that spatial relationship is not considered in such supervision since all entries are treated equally. To compensate for this lack of spatial supervision, we propose to additionally apply spatial softmax-based loss upon assignment matrix.

Based on aforementioned observations, this paper introduces AffineFormer, a novel cross-view Transformer equipped with geometric-aware deformable local attention and selective global-local message fusion, supervised by hybrid spatial softmax/classification-based loss. Following previous work~\cite{chen2022aspanformer}, we regress intermediate flow map during attention process, from which an affine-based deformation field is estimated. This deformation field is then utilized to project the local attention span from the source view to the target view. In parallel with local attention, we also incorporate coarse global attention, where the local and global messages are fused based on the reliability of the flow. As supervision, we apply a  Extensive experiments on both two-view pose estimation and visual localization demonstrate the effectiveness of our method.

\section{Related Works}
\subsection{Global-Local Attention for Image Matching}
Employing Transformer in dense/semi-dense matching boosts the matching capability of original features extracted from a single view, as has been studied by previous works~\cite{sun2021loftr,tang2022quadtree,chen2022aspanformer,wang2022matchformer,astr,casmtr}, yet the quadratic complexity of vanilla Transformer imposes challenges. Recently, some researchers~\cite{tang2022quadtree,chen2022aspanformer,astr} propose to combine coarse global attention and sparse local attention to handle high-resolution input while maintaining modest computation costs. Specifically, global attention establishes correlation across source and target tokens to guide local attention. In QuadTree Attention~\cite{tang2022quadtree}, full token sets are gradually tailored into different groups where sparse attention is performed only at more related target tokens at a fine level. AspanFormer~\cite{chen2022aspanformer} regresses intermediate flow maps with uncertainty to adaptively adjust attention span. ASTR~\cite{yu2023adaptive} utilizes neighborhood around intermediate matching points to generate local attention region for better local consistency.  

Sharing a similar practice with ASpanFormer, we regress intermediate flow field during cross attention process. However, instead of relying on learned uncertainty to determine attention pattern, we base our method on a geometric ground, where piece-wise affine deformation field are estimated to shape local attention span. 

\subsection{Deformable Attention}
In the context of the general Vision Transformer~\cite{dosovitskiy2020image}, deformable attention~\cite{xia2022vision,zhu2020deformable} is introduced as an augmentation to the vanilla attention framework. It dynamically shapes attention span based on local features, similar to the concept of deformable convolution in CNNs~\cite{dai16rfcn,dai17dcn}. Deformable attention predicts a set of offsets from sampled tokens to modify the sampling position. While it has shown effectiveness in various applications, plain deformable attention is not directly applicable to cross-view attention in image matching tasks. Additionally, the current deformable attention approach follows a fully data-driven method, making it challenging to interpret the learned deformation clearly. In contrast to the free-form deformable attention, we propose a geometry-driven deformable operation that explicitly models local deformation introduced by view changes, as shown in Fig.~\ref{deform}.

\subsection{Local Deformation Estimation}
Estimating local deformation patterns is a highly focused subject in the field of image matching. In the pre-deep learning era, traditional descriptors relied on manually designed shape detectors to guide the generation of local patches~\cite{rootsift,sift}. With rapid advancements of deep learning techniques, numerous studies have explored learning-based approaches for estimating deformations. OriNet~\cite{yi2016learning} and LIFT~\cite{yi2016lift} proposed to learn a canonical orientation for feature points, AffNet~\cite{mishkin2018repeatability} predicts additional affine parameters to enhance modeling capabilities. UCN~\cite{choy2016universal} and LF-Net~\cite{yi2016lift} take images as input and apply spatial transformation networks to intermediate features. Some CNN-based local features~\cite{luo2020aslfeat, alike, aliked} employs deformable convolution to generate dense deformation fields.

Unlike the aforementioned works that focus solely on single-view feature descriptions, we embed local shape estimation into a cross-view attention framework in a more principled manner.

\section{Methodology}
\begin{figure*}
    \centering 
	\includegraphics[width=0.9\textwidth]{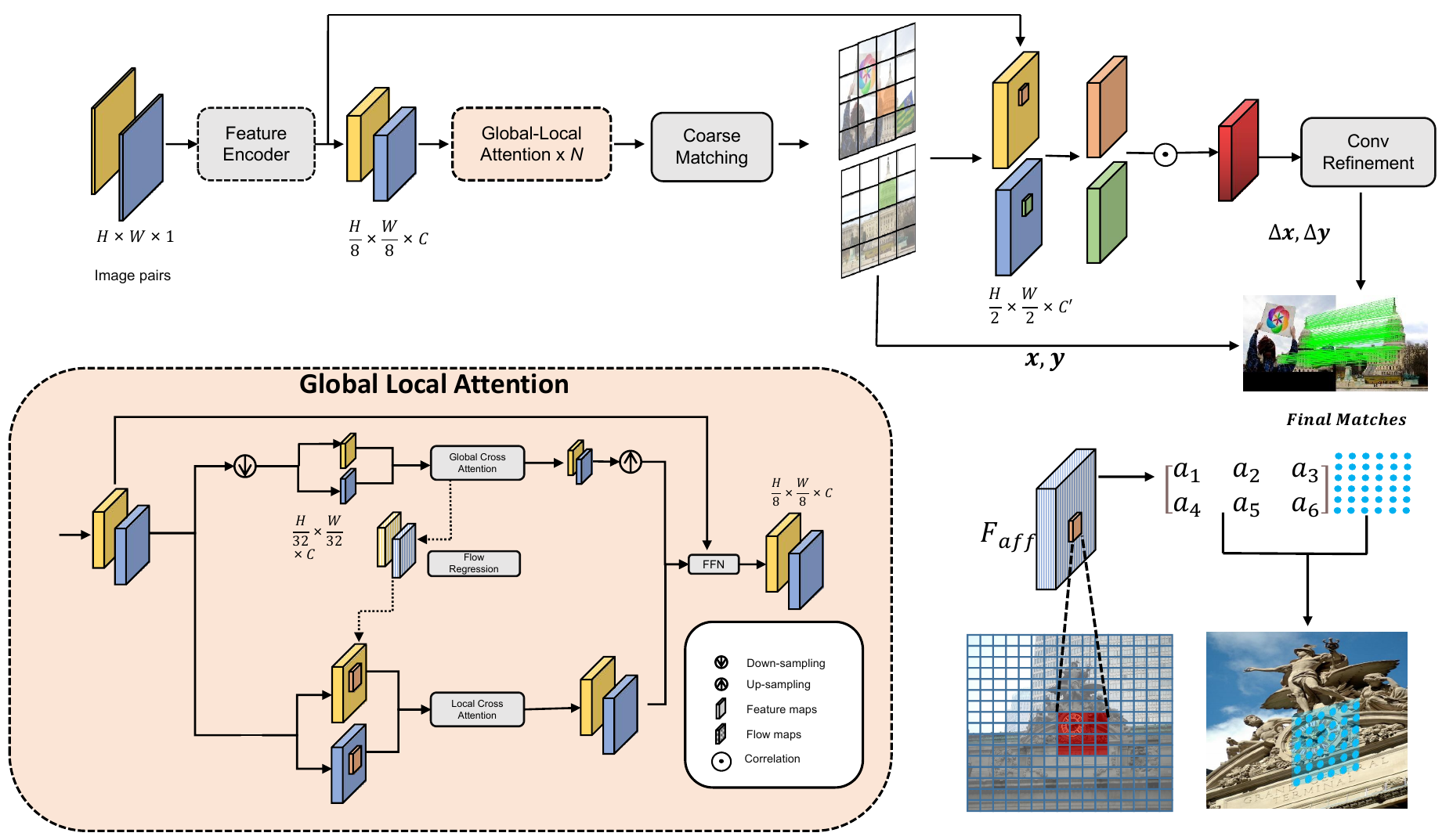}
	\caption{The overall structure of our proposed network. The network adopts iterative global-local attention operations to pass cross-view messages at both global and local scales. After identifying coarse level matches at $1/8$ resolution, a convolution refiner follows to predict correspondence residuals.}
	\label{fig:netarch}
\end{figure*}

In Fig.~\ref{fig:netarch}, we provide an overview of our network architecture, which inherits the paradigm of semi-dense matching~\cite{sun2021loftr}. Taking an image pair $I_A,I_B$ as input, our network generates coarse correspondences and then refine them. The nework begins with a CNN-based encoder to extract initial features in $\frac{1}{8}$ resolution for both images. These features are position-encoded through element-wise summation of sinusodinal signals~\cite{transformer} and passed through iterative self/cross attention blocks for enhancing. All self attention blocks are conducted at $\frac{1}{32}$ scale. For each cross attention, global attention is conducted at $\frac{1}{32}$ resolution, while affine-based local deformable attention is conducted at $\frac{1}{8}$ resolution. Local and global messages are then fused based on uncertainty of intermediate flow estimation to suppress unreliable local information. Self attention is also conducted at $\frac{1}{32}$ resolution between two cross attention blocks. 

\subsection{Global Attention}
In this part, we introduce global attention at $\frac{1}{32}$ resolution, which is used in self attention blocks and global branch in cross attention blocks.Formally, the input source/target features $F_s, F_t$ at $\frac{1}{8}$ scale are downsampled to $\frac{1}{32}$ through strided 2 average pooling. Vanilla dot-product attention is then performed to retrieve coarse message $m_c$, which is bilinear upsampled back to $\frac{1}{8}$ and fused with $F_s$ through a feed-forward network (FFN).:
\begin{align}
    F_s' &= \textbf{Avgpool}_{2\times2}(F_s) \\
    F_t' &= \textbf{Avgpool}_{2\times2}(F_t) \\
    m'&= \textbf{Attn}(W_qF_s',W_kF_t',W_vF_t')\\
    m&=\textbf{Up}_{2\times2}(m')\\
    \hat{F}_s&=F_s+\textbf{FFN}(F_s,m)\\
    &=F_s+\textbf{LN}(\textbf{DWConv}(F_s+\textbf{MLP}(m))
\end{align}
Here, $W_{(q/k/v)}$ denotes the linear transformation matrix for query/key/values vectors, \textbf{DWConv} means depth-wise convolution, \textbf{LN} means layer normalization. $\hat{F}_s$ is the updated source features, which is fed into following attention layers.

Note that global $msg$ will additionally be combined with local message in cross attention blocks, which will be introduced in the next section. We also would like to mention that even without local branch in cross attention, this basic global attention 
blocks has been a very competitive baseline, which is validated both in our experiment in Sec. 4.3 and a concurrent LoFTR's follow up work~\cite{eloftr}.

\subsection{Local Deformable Attention}
\label{sec:ga}
Global attention ensures long-range dependency, yet inevitably lost fine-level information due to lack of concentration and downsampling. To alleviate this issue, previous works~\cite{chen2022aspanformer,astr,tang2022quadtree} propose to enhance global attention with parallel fine-level attention. However, 
these works either adopts irregular sparse sampling or data-driven rectangular sampling, neglecting the importance to align local deformation across two-views.
Further more, fixed learned prior are used to fuse global and local message, yet not all local message are equally reliable due to inherent uncertainty. 

To address these issues, we propose affine-based deformable attention and selective message fusion. Concretely, we estimate intermediate deformation filed to align the focusing region for each token group. The retrieved message in local attention are further fused with global message based on estimated uncertainty. In the following part, we introduce the workflow and insights of this operation.

\subsubsection{Intermediate Flow Regression}
As mentioned in Sec. 3.1, global cross attention is conducted parallel with each local attention, which outputs coarse retrieved message $m_c \in R^{\frac{H}{32} \times \frac{W}{32} \times C}$ and attention matrix $A \in R^{N\times M \times H}$. Here $N,M$ denotes the number of source and target tokens, and $H$ denotes the head number in multi-head attention. Naturally, the attention matrix reflects the similarity between cross-view features and thus can be used to decode a rough intermediate flow map. Inspired by the global decoder in dense matching methods~\cite{dkm,roma}, we utilize weighted sum of position embeddings and a decoder to regress intermediate flows. Formally, mean of $A$ along the head dimension is computed as $\hat{A} \in R^{N\times M}$, which is used to aggregate weighted positional embeddings $P_t \in R^{M \times D}$ from the target image. Here, $D$ is the positional embeddings' channel number. A convolution-based decoder is followed to regress flows and uncertainty:
\begin{align}
    \Phi=\textbf{Conv}(\hat{A}P), ~~\Phi \in R^{\frac{H}{32} \times \frac{W}{32} \times 4}.
\end{align}
Each element $\phi_{i}=[u_{xi},u_{yi},\sigma_{xi},\sigma_{yi}]$ from $\Phi$ indicates the corresponding flow coordinates $u_{xi},u_{yi}$ and uncertainty $\sigma_{xi},\sigma_{yi}$ in each location. Drawing inspiration from previous works~\cite{chen2022aspanformer,kfnet,pdcnet+} that adopts a probabilistic framework to model flow uncertainty, we take uncertainty $\sigma_{xi},\sigma_{yi}$ as stand deviation in a two-dimensional Gaussian distribution and train them in a self-supervised manner. The estimated flow map is bilinear upsampled to $\frac{1}{8}$ resolution and is used to estimate patch-wise affine field.

\subsubsection{Affine Field Estimation}
Flow maps recovered by coarse attention matrix roughly reflect corresponding regions for each location in the source feature map, yet flow in free-form is inevitably corrupted by outliers and doesn't reflect priors in two-view geometry. For example, point correspondences from a local area without large depth fluctuation can be well approximated by an affine transformation. 

To embed geometric priors into deformation field, we estimate piece-wise affine parameters from intermediate flow map $\Phi$. Concretely, source flow map $\Phi$ is grouped by non-overlapping windows with size $l$. For each $l \times l$ window, a set of affine parameters is estimated from the local flow. Formally, we denote coordinates of each point $i$ in a local window as $[x_i,y_i]$, while its corresponding flow as $[\hat{x}_i,\hat{y}_i]$. The affine to be estimated is denoted as $A \in R^{2 \times 3}$. We set the last column of $A$ as difference between mean of  $[x_i,y_i]$, $[\hat{x}_i,\hat{y}_i]$,
\vspace{-0.5em}
\begin{align}
\begin{bmatrix}
    a_{13} \\
    a_{23}      
\end{bmatrix}
=\frac{1}{N}\sum_{i=1}^N
\begin{bmatrix}
    x_i-\hat{x}_i\\
    y_i-\hat{y}_i    
\end{bmatrix}
\end{align}
The rest of elements in $A$ are estimated through linear least squares,
\vspace{-0.5em}
\begin{align}
 \begin{bmatrix}
    a_{11} & a_{12}\\
    a_{21} & a_{22}   
\end{bmatrix}^T
=(C^TC)^{-1}C^T
\begin{bmatrix}
    \hat{x}_1' & \hat{y}_1'\\
     \hat{x}_2' & \hat{y}_2'\\
     \hdotsfor{2} \\
     \hat{x}_N' & \hat{y}_N'
\end{bmatrix}
\end{align}

where
\begin{align}
    C=\begin{bmatrix}
    x_1' & y_1'\\
     x_2' & y_2'\\
     \hdotsfor{2} \\
     x_N' & y_N'
    \end{bmatrix}\\
\end{align}
and
\begin{align}
    x_i'=x_i-a_{13}, y_i'=y_i-a_{23}\\
    \hat{x}_i'=\hat{y}_i-a_{13}, \hat{y}_i'=\hat{y}_i-a_{23}
\end{align}

To reduce misalignment caused by noisy flow, we further regularize the estimated $A$ through a series of operations, including constraining the underlying  scale, rotation and shearing. More details about affine regularization are provided in supplementary materials.

As a result, the flow field $\Phi \in R^{\frac{H}{8} \times \frac{W}{8} \times 4}$ is converted to affine parameters field $F_{aff}  \times R^{\frac{H}{8l} \times \frac{W}{8l} \times 6}$.

\subsubsection{Deformable Attention}
Ideally, the source/target patch in local attention should be aligned to capture the most relevant features. However, achieving this alignment with a rectangular attention pattern is often infeasible due to complex local deformations.

To address this issue, we leverage the estimated affine field to sample a deformed target patch for each source patch. Concretely, we reuse the non-overlapping windows with size $l$ in previous affine field estimation as source patches. For each source patch $S$, we use the corresponding affine parameters $f_{aff} \in R^6$ from $F_{aff}$ to project an affine patch $\hat{S}$ in target feature map. To ensure better coverage, the size of target patch is set as $\alpha l$, which is $\alpha$ times larger than the source token patch. Illustration of this process can be seen Fig.~\ref{fig:netarch}.

Given query and key/value feature maps $F_q,F_k/v$, we sample tokens uniformly in $S,\hat{S}$, which yields tokens $f_q \in R^{l^2 \times D}, f_k/v \in R^{\alpha^2l^2 \times D}$ for each patch. Local attention is performed within each patch pair to generate local message.

\subsection{Selective Message Fusion}
\label{sec:adaptive}
\begin{figure*}
    \centering 
	\includegraphics[width=0.9\textwidth]{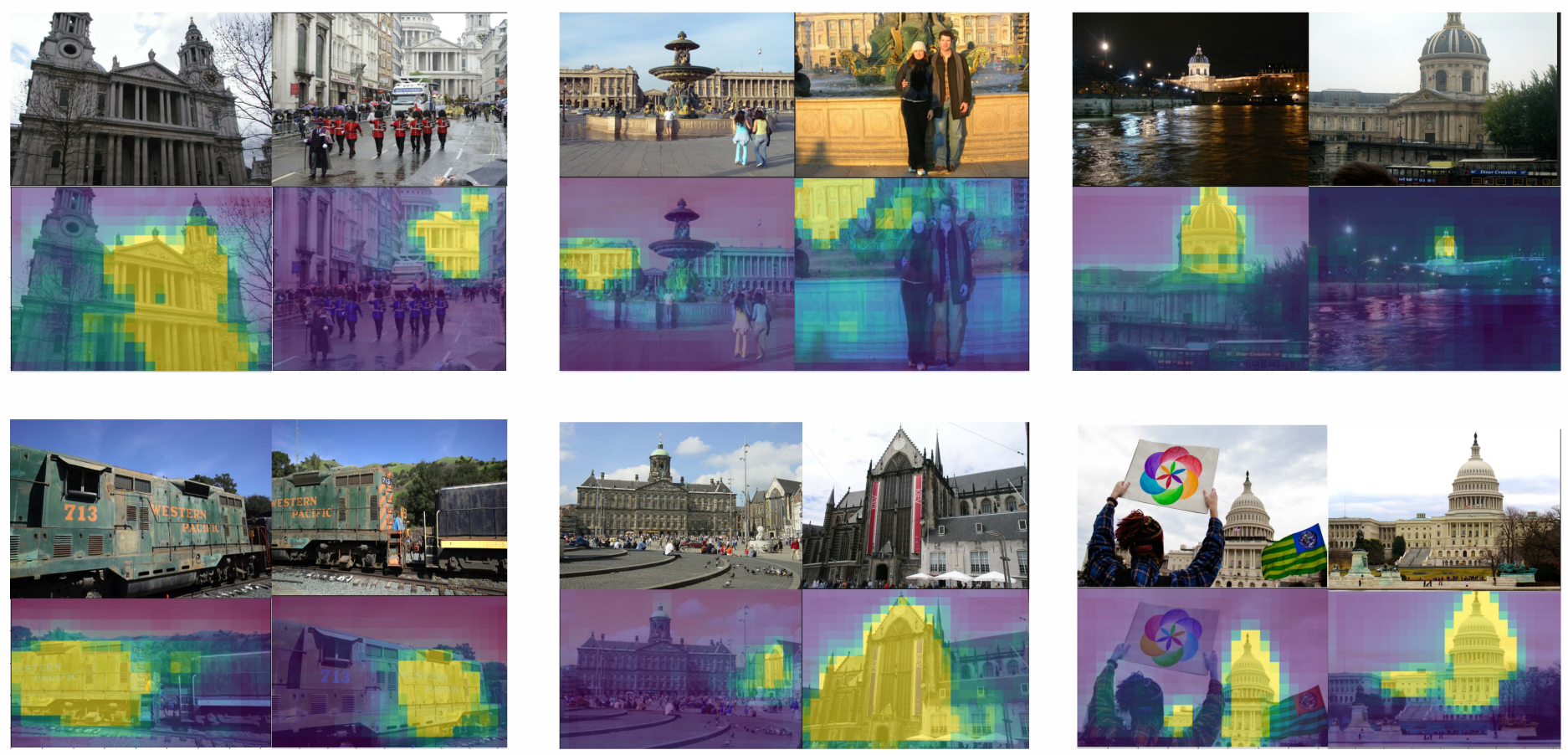}
	\caption{Heatmap for selective local fusion score.}
	\label{score}
 \vspace{-1em}
\end{figure*}

Simply averaging or concatenating global and local message $m_g,m_l$ is a straightforward way for message fusion, which, however, is problematic since $m_l$ may come from inaccurate flow estimation or non-overlapping regions. To address this issue, we use predicted uncertainty to weight local message.
Concretely, for each position in the source feature map, we index the corresponding uncertainty $\sigma_x,\sigma_y$ from the intermediate flow map. The fused message is calculated as a weighted sum of $m_g,m_l$:
\begin{align}
    m&=p_1m_g+p_2m_l,\\
    [p_1,p_2]=\textbf{softmax}&(\alpha,\beta[(1+\gamma \textbf{ReLu}(\sigma_x+\sigma_y)]^{-1})
\end{align}
$\alpha,\beta$ are learnable parameters to balance prior weight for global and local messages, which are modified by uncertainty-related factor. A learnable parameter $\gamma$ controls its sensitivity. Through this formulation, large uncertainty results in lower weight in message fusion. The obtained fuse message $m$ is used to update source features through a feed-forward network as introduced in Sec. 3.1. An illustration of learned fusion heatmap can be seen in Fig~\ref{score}.
The produced score map helps our network to sharply focus on co-visible and salient regions, discarding the non-relevant areas in message fusion.

\subsection{Match Determination}

After all attentional blocks, the enhanced features $\tilde{F}_A \in R^{n\times c}, \tilde{F}_B \in R^{m \times c}$ are fisrt used to generate correlation matrix $C=\tau \tilde{F}_A \tilde{F}^T_B \in R^{n \times m}$, where $\tau$ is a temperature parameter,  followed by dual-direction softmax operation to produce assignment matrix $S$. We retain coarse-level correspondences $M_c$ by mutual nearest neighbor (MNN) and threshold of 0.2 on dual-softmax score. 

To refine coarse matches $M_c$ are in $1/8$ resolution, a local correlation-based refinement block is adopted. For each coarse match, we sample a local window with size $w$ from the source and target feature map in $1/2$ resolution, which yield local patches $p_s,p_t \in  R^{w \times w \times D}$. For feature in source patch $p_s$, we calculate its correlation score with target patch $p_t$, which are flattened into correlation feature and fed into a convolutional refiner to predict fine level residuals conditioned on coarse match. 

\begin{figure*}
    \centering 
	\includegraphics[width=0.85\textwidth]{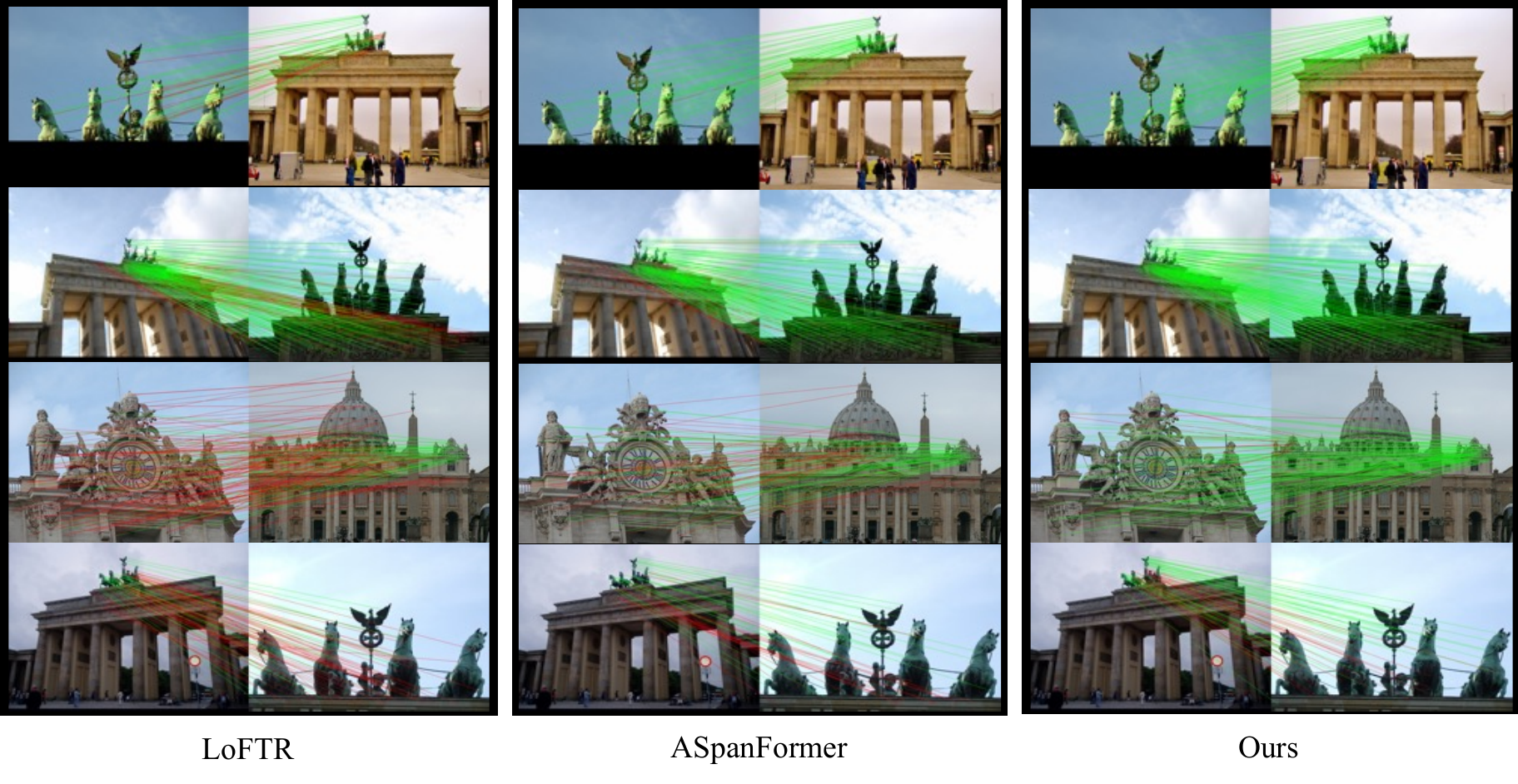}
	\caption{Visualization of image matches from LoFTR~\cite{sun2021loftr}, ASpanFormer~\cite{chen2022aspanformer} and our method, where \textcolor{green}{green} lines represent inlier matches, while \textcolor{red}{red} lines represent outlier matches.}
	\label{match}
\end{figure*}

\subsection{Loss Formulation}

Loss of our method consists of three parts, (1) coarse-level loss, (2) fine-level loss, (3) flow estimation loss.

Coarse level loss includes two terms, the classification loss $L_{ce}$ and spatial softmax loss $L_{cs}$. In cogruent with previous works, we re-reproject points in each image pair using ground-truth depth and camera poses, where the mutual nearest match $M_{gt}$ are considered as ground truth match. The classification loss $L_{ce}$ is defined as a focal loss using assignment matrix $S$:
\begin{align}
    L_{ce}=&-\sum_{(i,j)\in M_{gt}} (1-S(i,j))^\gamma \text{log}(S(i,j)) \nonumber\\
    &-\sum_{(i,j)\notin  M_{gt}} S(i,j)^\gamma\text{log}(1-S(i,j)).
\end{align}
One limitation of classification loss is that all mismatches suffer the same loss penalty no matter how far they are from the ground truth. Taking inspiration from previous works on learned descriptor~\cite{caps}, we adopt an additional spatial softmax loss as compensation, where close mismatch should produce lower loss than 'distant' ones.
\begin{align}
    L_{cs}=\frac{1}{|M_{gt}|}\sum_{i \in M_{gt}[:,0]} [\sum_jS(i,j)P_{ij}-P_{i}^{gt}]^2
\end{align}

Here, $P_{ij}$ denotes coordinates for each entry in the assignment matrix, while $P_{i}^{gt}$ denotes the ground truth match coordinate for left point $i$. Spatial softmax loss encourage maximization of positive entries in assignment matrix, which is in the same direction of classification loss. Additionally, it also considers spatial relationship and penalizes mismatch according to their distance to the ground truth. In our ablation study in Sec. 4.3, we fine the simple modification on loss term largely benefit overall performance. 

For intermediate flow, we follow previous works~\cite{chen2022aspanformer,kfnet} to minimize the log-likelihood for the estimated Gaussian distribution. Formally, given flow estimation $\Phi$ from each layer and ground truth flow $D^{gt}$, $L_{flow}$ is defined as:
\begin{align}
    L_{flow}=-\frac{1}{|D^{gt}|}\sum_{ij} \text{log}(P(D^{gt}_{ij}|\Phi_{ij})).
\end{align}
More details about flow regression are provided in the supplementary material.

The fine-level loss is defined as the l2-distance between regress refine offset and ground truth. 

In summary, we formulate the loss as:
\begin{align}
L= L_{ce} + L_f+ \lambda_1L_{cs}  +\lambda_2 L_{flow}.
\end{align}

\subsection{Implementation Details}
Our network uses ResNet-18~\cite{resnet} CNN backbone. We use 4 interleaved self/cross attention blocks for feature update. For local cross attention, we set window size $l=4$. For fine-level refinement, we set local window size $w=5$.
For supervision, we set $\lambda_1$ as 1 for outdoor model and 5 for indoor model, $\lambda_2$ is set as 0.1.

We train two different models for indoor and outdoor scenes respectively on ScanNet and Megadepth. Both models follows the same training scheme in previous works~\cite{sun2021loftr,chen2022aspanformer,tang2022quadtree}, which lasts for 30 epochs on 8 V-100 GPUs. More details of implementation and training are available in the supplementary material.

\let\saveFloatBarrier\FloatBarrier
\let\FloatBarrier\relax
\section{Experiments}
\let\FloatBarrier\saveFloatBarrier
\subsection{Two-view Pose Estimation}
\label{sec:two_view_eval}
\smallskip\noindent\textbf{Datasets.} We use ScanNet~\cite{dai2017scannet} and MegaDepth~\cite{li2018megadepth} datasets to evaluate the matching ability of our method in indoor scenes and outdoor scenes. We follow evaluation protocols to select 1500 image pairs from two datasets respectively, where the relative poses are recovered through OpenCV ransac, as is done in previous works~\cite{sun2021loftr,chen2022aspanformer,tang2022quadtree,astr}. For ScanNet, we resize all images to [640,480] resolutions. For MegaDepth, we resize all images to [1152,1152]. LoFTR(E) and TopicFM only reports outdoor trained model, thus the corresponding number for indoor evaluation are omitted.

\smallskip\noindent\textbf{Comparative methods.} We compare the proposed method with 1) sparse approaches, 2) semi-dense approaches that outputs $1/8$ resolution coarse matches with local refinement, including LoFTR~\cite{sun2021loftr}, QuadTree Attention~\cite{tang2022quadtree}, ASpanFormer~\cite{chen2022aspanformer}, TopicFM~\cite{topicfm} and efficient LoFTR~\cite{eloftr}. We also include a series fully dense methods producing dense warp for reference. 

\input{table/two_view}

\smallskip\noindent\textbf{Results.} As presented in Table~\ref{two-view}, our method outperforms all sparse and semi-dense methods on both indoor and outdoor dataset. We also report cross-dataset generalization results by evaluating MegaDepth model on ScanNet. Even without training on any indoor scenes, our outdoor model demonstrates high accuracy on indoor scenerios.

The speed-optimized efficient loftr demonstrates impressive performance with several advanced network designs, including rotary positional encoding, two stage refinement and RepVGG backbone, which can also be adopted to enhance our method (the reported results don't use these enhancements). 

\input{table/localization}

\subsection{Visual Localization}
\smallskip\noindent\textbf{Datasets.} Apart from two-view pose estimation, we further evaluate our network in the visual localization pipeline, where two stand benchmarks InLoc~\cite{taira2018inloc} and Aachen Day-Night v1.1~\cite{zhang2021reference,sattler2012imageBMCV, aachen} are used to demonstrate the performance in indoor and outdoor scenes. We embed our method to Hloc~\cite{sarlin2019coarse} pipeline for evaluation. We use the model trained on the MegaDepth to localize both inloc and Aachen datasets. All input images are resized so that the longest dimension is 1024.

\smallskip\noindent\textbf{Results.} The results are reported on Table~\ref{localization}. Our method demonstrates highest accuracy on Aachen dataset, and similar performance with a concurrent work efficient LoFTR~\cite{eloftr} on InLoc dataset. Noted that the improvements proposed in our method is orthogonal to efficient LoFTR. Generally, our method shows strong generalization ability in visual localization settings.

\subsection{Ablation Study}
We conducted an ablation study on the ScanNet dataset, following the protocol outlined in Section~\ref{sec:two_view_eval}. The results in shown in Table~\ref{ablation}. Our baseline uses 4 interleaved self/cross attention, where the cross attention is only conducted at $\frac{1}{32}$ scale without fine-level local attention. 

To study the effect of loss design, we test two settings: (1) replace focal loss with spatial softmax loss (row 2), (2) use spatial softmax loss as an additional term (row 3). We observe that direct replacement results in worse results, indicating focal loss is essential for learning distinctive features. Adding spatial softmax loss brings considerable improvement, reflecting the importance to enforce spatial smoothness in loss design.

We then sequentially add local attention with fixed-size rectangular span (row 4), add uncertainty-based selective fusion in Sec. 3.3 (row 5), and apply affine-based deformation (row 6). All proposed components make notable contributions over baseline, validating the effectiveness of our method designs.
\input{table/ablation}

\subsection{Efficiency Evaluation}
\label{sec:cost}
\input{table/cost}

In this section, we conduct a comparison of different semi-dense methods on their model size and inference cost. In addition to our normal setting, we also provide a light variant with very slim backbone (600k parameters) and 2 attention layers (denoted as ours-L). Details for the light setting can be found in supplementary materials. We apply the similar light-weight modifications to LoFTR (denoted as LoFTR-L). 

As can be seen in Tab.~\ref{cost}, our normal version network shares similar cost and model size with LoFTR while delivers significant performance gain. Our light version network uses only 15\% flops and 18\% parameters to reach LoFTR's performance, while reducting LoFTR to the same level results in largely degenerated performance. Overall, our method shows good performance under different cost budget.

\section{Conclusion}
In this paper, we introduce AffineFormer, a novel semi-dense matcher equipped with affine-based deformable local attention and selective message fusion. We capture local deformation caused by viewpoint changes through the estimation of affine transformation field, which is used to shape local attention patterns. We then propose to fuse global-local message robustly through adaptive fusion. The effectiveness of spatial softmax-based loss is also studied, which is neglected in previous works. Extensive experiments demonstrates our method's effectiveness in geometry estimation.

\clearpage

{\small
\bibliographystyle{ieee_fullname}
\bibliography{egbib}
}

\clearpage

\end{document}


\section*{Supplementary}

\section{Implementation Details}

We provide addition implementation details in this part, including auxiliary flow regression, affine estimation and general training settings.

\subsection{Intermediate Flow Regression}
We inherit formulation of auxiliary flow regression from ASpanFormer~\cite{chen2022aspanformer}, where the flows are treated as random variables in 2D gaussian distribution. Concretely, for a pixel in source image, the probability that it correspond to $(x,y)$ in target image is given by 

\begin{align}
P(x,y)= \frac{1}{2\pi\sigma_x\sigma_y}\text{exp}(-\frac{(x-u_x)^2}{2{\sigma_x}^2}-\frac{(y-u_y)^2}{2{\sigma_y}^2})
\end{align}

The mean $u_x,u_y$ and standard deviation $\sigma_x,\sigma_y$ is predicted by a network introduced in main paper Sec. 3.2.1. During training, we use log liklihood loss to supervise flow prediction. More specifically, for each pixel the flow loss is given by

\begin{align}
    L_{flow}&=log [\frac{1}{2\pi\sigma_x\sigma_y}\text{exp}(-{\frac{(x_{gt}-u_x)^2}{2\sigma_x^2}-\frac{(y_{gt}-u_y)^2}{2\sigma_y^2}})]\\
    &= log 2\pi+log\sigma_x+log\sigma_y+\frac{(x_{gt}-u_x)^2}{2\sigma_x^2}+\frac{(y_{gt}-u_y)^2}{2\sigma_y^2}
\end{align}

where $[x_{gt},y_{gt}]$ is ground truth correspondences coordinates. After Denoting $w_x=log\sigma_x,w_y=log\sigma_y$ and dropping $log2\pi$, then

\begin{align}
L_{flow}&= w_x+w_y \nonumber\\
&+\frac{1}{2}e^{-2w_x}(x_{gt}-u_x)^2+\frac{1}{2}e^{-2w_y}(y_{gt}-u_y)^2
\end{align}

First two terms encourage the network to decrease prediction uncertainty. Last two terms use inverse exponential to weight l2-distance between ground truth and predicted coordinates, which encourages prediction with lower accuracy to have a higher standard deviation (or lower confidence). As a whole, the flow loss encourages lower flow estimation error and adjust uncertainty accordingly.

\subsection{Regularizing Affine Estimation}
As mentioned in main paper, the accuracy of estimated affine matrices can be affected by noisy intermediate flow. To alleviate this issue, we further apply regularization upon each affine matrix. 

Concretely, we pick top 50\% points w.r.t. flow uncertainty within each local patch to estimate affine. The estimated $A$ will be decomposed into scale, rotation and shearing factor as following,

\begin{align}
\begin{bmatrix}
    a_{11} & a_{12} \\
    a_{21} & a_{22}     
\end{bmatrix}
=
\begin{bmatrix}
    cos\theta & -sin\theta\\
    sin\theta & cos\theta     
\end{bmatrix}
\begin{bmatrix}
    1 & m\\
    0 & 1     
\end{bmatrix}
\begin{bmatrix}
    s_x & 0\\
    0 & s_y     
\end{bmatrix}
\end{align}

which yields, 

\begin{align}
    \theta &=\textbf{arctan}(\frac{a_{21}}{a_{11}})\\
    s_x &=\sqrt{(a_{11}^2+a_{21}^2)} \\
    s_y &=|a_{22}cos\theta-a_{12}sin\theta| \\
    m&=\frac{a_{12}cos\theta+a_{22}sin\theta}{s_y}
\end{align}

We constrain the scale $s_x, s_y$ to be in range of [0.5, 4], the shearing factor $m$ to be in range [-0.5, 0.5] and the rotation angle $\theta$ to be in range [$-\frac{\pi}{3},\frac{\pi}{3}$]. We further deprecate local message from border grids (setting the corresponding selective fusion score as 0). In Tab.~\ref{regularization}, we provide ablation on affine estimation w. and w/o regularization.

\input{table/reg}

 
\subsection{Training Settings}
We inherit data splits from LoFTR~\cite{sun2021loftr} and its following works~\cite{chen2022aspanformer,tang2022quadtree,astr} to train our model on both ScanNet~\cite{scannet} and MegaDepth~\cite{mega}. ScanNet model is trained with batch size 32 and initial learning rate $3e^{-3}$, where a linear warm-up us applied for the first epoch. Learning rate is halved at epoch [3, 6, 9, 12, 15, 18, 21, 24, 27]. Megadepth model is trained with batch size 8 and initial learning rate $1e^{-3}$, where a linear warm-up us applied for the first 3 epoches. Learning rate is halved at epoch [8,12,16,20,24].

\subsection{Light-weight variant}

\begin{figure*}
    \centering 
	\includegraphics[width=1\textwidth]{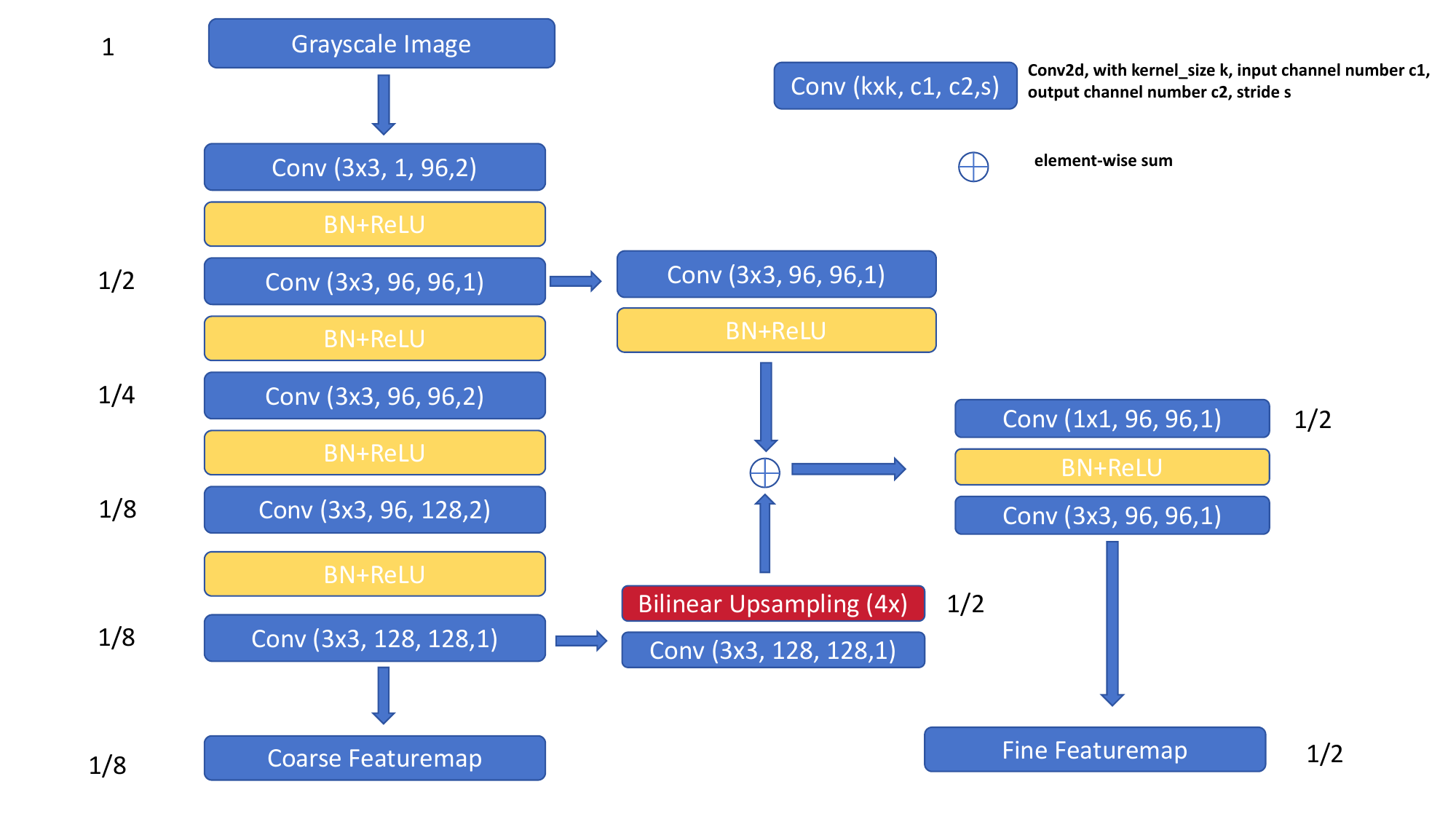}
	\caption{Network structure for backbone of light-weight variant. The backbone takes a full scale grayscale image as input and outputs coarse/fine feature map in $\frac{1}{8},\frac{1}{2}$ respectively. The coarse feature is fed into our attention-based cross-view network for feature update, while the fine level network is used for match coordinate refinement.}
	\label{cnn_comparison}
\end{figure*}

We reduce our full model on 3 points to obtain the light-weight version: (1) channel number for all attention layers is cut from 256 to 128, (2) number of interleaved attention blocks is cut from 4 to 2, (3) CNN backbone is reduced from ResNet-18 to a light vgg style network, of which the structure can be seen in Fig~\ref{cnn_comparison}.


\section{More Visualizations}
\begin{figure*}
    \centering 
	\includegraphics[width=1\textwidth]{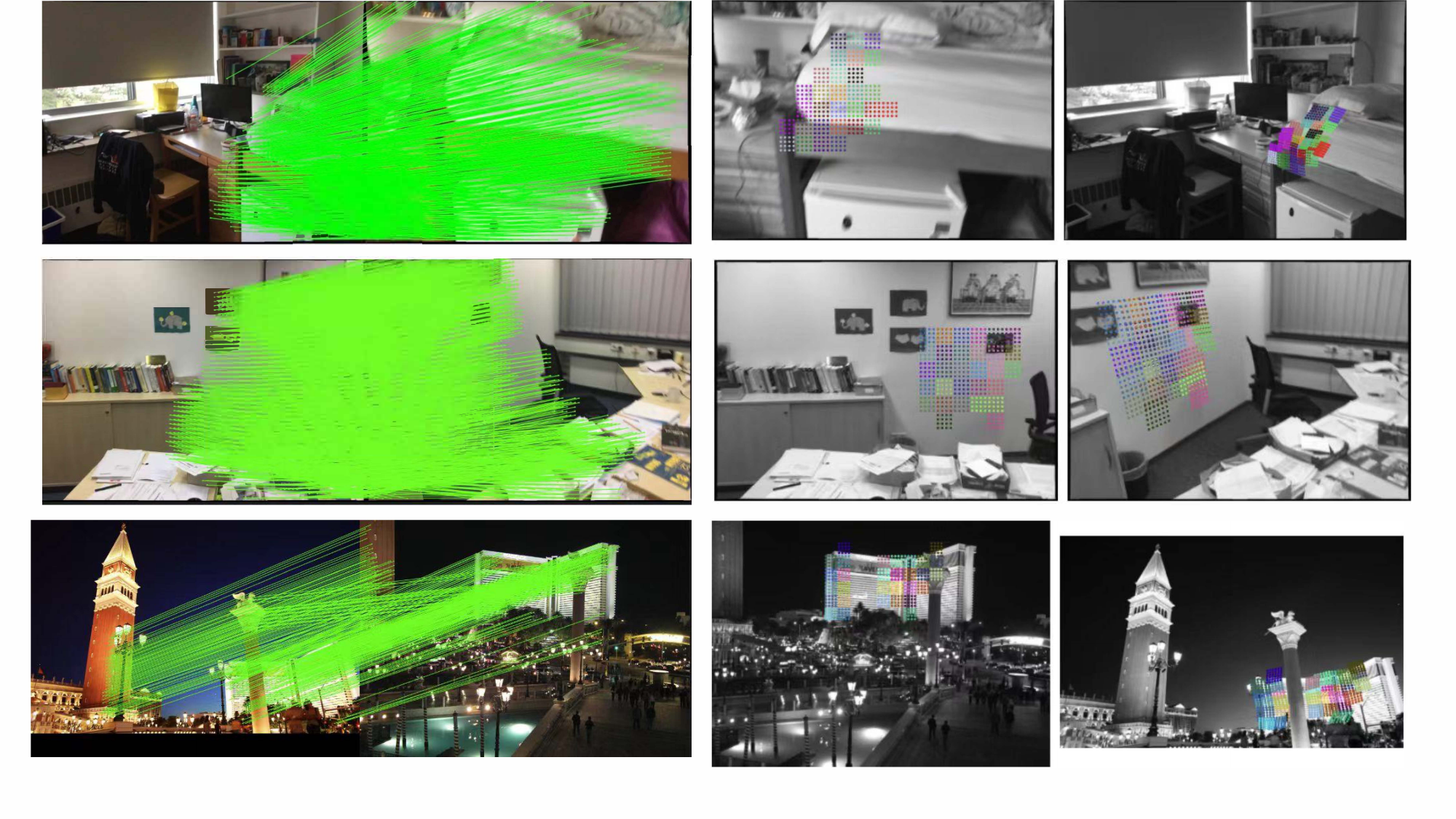}
	\caption{Matching results and estimated affine filed (top 20 according to uncertainty).}
	\label{addvis}
\end{figure*}

We provide additional visualizations on matches and affine attention span in Fig.~\ref{addvis}

\section{Limitations and Future Works}

For a fair comparison with previous works, our network is based on a relatively old fashion, including ResNet backbone and absolute sinusoidal positional encoding, which can be replaced by advanced feature extractor~\cite{eloftr,roma} and rotary positional encoding~\cite{eloftr}. Furthermore, although our method is effective in general scenarios, we find the affine-based deformation may not hold in some special settings, such as non-rigid scenes or objects with very fine-grained geometry. Potential future works include modernize our network with recent progress in backbone/Transformer designs and further improve the robustness of local deformation modeling.  

{\small
\bibliographystyle{ieee_fullname}
\bibliographystyle{unsrt}
\bibliography{egbib}
}

%% file: preamble.tex
\usepackage[dvipsnames]{xcolor}


%% file: table/two_view.tex
\begin{table*}
\centering
\resizebox{0.8\textwidth}{!}{
\begin{tabular}{cccc||ccc}
\Xhline{1pt}
\noalign{\smallskip}
\multirow{2}{*}{\textbf{Method}} & \multicolumn{3}{c}{\textbf{ScanNet}} & \multicolumn{3}{c}{\textbf{MegaDepth}} \\
\noalign{\smallskip}
\cline{2-7}
\noalign{\smallskip}
 & AUC@$5\degree$ & AUC@$10\degree$ & AUC@$20\degree$ & AUC@$5\degree$ & AUC@$10\degree$ & AUC@$20\degree$ \\
\noalign{\smallskip}
\Xhline{1pt}
\noalign{\smallskip}
\textit{SP~\cite{detone2018superpoint}+SuperGlue~\cite{superglue}} & 16.2 & 33.8 & 51.8 & 49.7 & 67.1 & 80.6 \\
\textit{SP~\cite{detone2018superpoint}+LightGlue~\cite{lightglue}} & 14.8 & 30.8 & 47.5 & 49.9 & 67.0 & 80.1 \\
\Xhline{1pt}
\textit{LoFTR~\cite{sun2021loftr}} & 22.0 (16.9) & 40.8 (33.6) & 57.6 (50.6) & 52.8 & 69.2 & 81.2 \\
\textit{QuadTree~\cite{tang2022quadtree}} & 24.9 (19.0) & 44.7 (37.3) & 61.8 (53.5) & 54.6 & 70.5 & 82.2 \\
\textit{MatchFormer~\cite{wang2022matchformer}} & 24.3 (15.8) & 43.9 (32.0) & 61.4 (48.0) & 53.3 & 69.7 & 81.8 \\
\textit{ASpanFormer~\cite{chen2022aspanformer}} & 25.6 (19.6) & 46.0 (37.7) & 63.3 (54.4) & 55.3 & 71.5 & 83.1 \\
\textit{TopicFM~\cite{topicfm}} & - (17.3) & - (35.5) & - (50.9) & 54.1 & 70.1 & 81.6 \\
\textit{LoFTR (E)~\cite{eloftr}} & - (19.2) & - (37.0) & - (53.6) & 56.4 & 72.2 & 83.5 \\
\textit{\textbf{Ours}} & \textbf{27.1} (\textbf{22.0}) & \textbf{47.5} (\textbf{40.9}) & \textbf{64.8} (\textbf{58.0}) & \textbf{57.3} & \textbf{72.8} & \textbf{84.0} \\
\Xhline{1pt}
\textit{PDCNet+(H)~\cite{pdcnet+}} & 20.3 & 39.4 & 57.1 & 51.5 & 67.2 & 78.5 \\
\textit{CasMTR~\cite{casmtr}} & 27.1 & 47.0 & 64.4 & 59.1 & 74.3 & 84.8 \\
\textit{DKM~\cite{dkm}} & 29.4 & 50.7 & 68.3 & 60.4 & 74.9 & 85.1 \\
\textit{RoMa~\cite{roma}} & 31.8 & 53.4 & 70.9 & 62.6 & 76.7 & 86.3 \\
\Xhline{1pt}
\end{tabular}
}
\caption{Two-view pose estimation results on ScanNet dataset~\cite{dai2017scannet} in indoor scenes and MegaDepth dataset~\cite{mega} in outdoor scenes. Figures in bracket are results of evaluating outdoor-trained model on ScanNet dataset.}
\label{two-view}
\end{table*}

%% file: table/localization.tex
\begin{table*}
\centering
\resizebox{0.95\textwidth}{!}{ 
\begin{tabular}{c c c c||c c c}
\Xhline{1pt}
\noalign{\smallskip}
\multirow{2}{*}{\textbf{Method}}  & \textbf{DUC1} & \textbf{DUC2} & \multirow{2}{*}{\textbf{Mean}}& \textbf{Day} & \textbf{Night} & \multirow{2}{*}{\textbf{Mean}}\\
\noalign{\smallskip}
\cline{2-3} \cline{5-6}
\noalign{\smallskip}
 & \multicolumn{2}{c}{(0.25m,2$\degree$) / (0.5m,5$\degree$) / (1m,10$\degree$)} \\
\noalign{\smallskip}
\Xhline{1pt}
\noalign{\smallskip}
\textit{SP~\cite{sp}+SuperGlue~\cite{superglue}}  & 47.0 / 69.2 / 79.8 &  53.4 / 77.1 / 80.9 & 67.90 & 89.8 / 96.1 / \textbf{99.4} & 77.0 / 90.6 / \textbf{100.0} &  92.15  \\
\textit{SP~\cite{sp}+LightGlue~\cite{lightglue}}  & 49.0 / 68.7 / 80.8 &  55.0 / 74.8 / 79.4 & 67.95 & \textbf{90.2} / 96.0 / \textbf{99.4} & 77.0 / 91.1 / \textbf{100.0} &  92.28  \\
\hline
\textit{LoFTR~\cite{sun2021loftr}} & 47.5 / 72.2 / 84.8 & 54.2 / 74.8 / 85.5 & 69.83 & 88.7 / 95.6 / 99.0 & 78.5  / 90.6 / 99.0 & 91.90  \\
\textit{MatchFormer~\cite{wang2022matchformer}} & 46.5 / 73.2/ 85.9 & 55.7 /71.8 / 81.7 & 69.13 & - /- / - & - / - / - & - \\
\textit{ASpanFormer~\cite{chen2022aspanformer}} & 51.5 / 73.7 / 86.4 & 55.0 / 74.0 / 81.7 & 70.38 & 89.4 / 95.6 / 99.0 & 77.5 / 91.6 / 99.5 & 92.10 \\
\textit{PATS~\cite{pats}} & 55.6 / 71.2 / 81.0 & \textbf{58.8} / \textbf{80.9} / 85.5  & 71.45 & 89.6 / 95.8 / 99.3 & 73.8 / \textbf{92.1} / 99.5 & 91.68\\
\textit{TopicFM~\cite{topicfm}} & 52.0 / 74.7 / \textbf{87.4}  & 53.4 / 74.8 / 83.2 & 72.17 & 90.2 / 95.9 / 98.9 & 77.5 / 91.1 / 99.5 & 92.18 \\
\textit{LoFTR (E)~\cite{eloftr}} & 52.0 / 74.7 / 86.9  & 58.0 / \textbf{80.9} / \textbf{89.3} & \textbf{73.63} & 89.6 / \textbf{96.2} / 99.0 & 77.0 / 91.1 / 99.5 & 92.06 \\
\textit{\textbf{Ours}} & \textbf{56.1} / 74.7 / 86.9 & 55.0 /79.4 / 87.0 &  73.18 & 89.9 / \textbf{96.2} / 98.9 & \textbf{79.1} / 91.1 / 99.5 & \textbf{92.45} \\\Xhline{1pt}
\end{tabular}
}
\caption{Visual localization results on InLoc~\cite{taira2018inloc} and Aachen Day-Night v1.1 dataset.}
\vspace{-0.5em}
\label{localization}
\end{table*}

%% file: table/ablation.tex
\begin{table}
\centering
\resizebox{0.5\textwidth}{!}{
\begin{tabular}{l>{\centering\arraybackslash}m{1.2cm}>{\centering\arraybackslash}m{1.2cm}>{\centering\arraybackslash}m{1.2cm}}
\Xhline{1pt}
\noalign{\smallskip}
\multirow{2}{*}{\textbf{Design}}  & \multicolumn{3}{c}{\textbf{Pose Estimation AUC}} \\
\noalign{\smallskip}
\cline{2-4}
\noalign{\smallskip}
& @$5\degree$ & @$10\degree$ & @$20\degree$ \\
\noalign{\smallskip}
\Xhline{1pt}
\noalign{\smallskip}
\textit{baseline} & 24.0 & 44.4 & 61.5  \\
~\textit{replace focal loss with s.s. loss} & 21.5 & 40.1 &  57.9 \\
~+\textit{s.s. loss} & 25.1 & 45.6 &  62.8 \\
~+\textit{local attn.} & 26.0 &  46.1 & 63.4 \\
~+\textit{selective fusion}& 26.6 &  46.8 & 63.9 \\
~+\textit{affine estimation.} & 27.1 & 47.5 & 64.8 \\
\Xhline{1pt}
\end{tabular}
}
\caption{Ablation study on ScanNet dataset~\cite{dai2017scannet}.}
\label{ablation}
\end{table}

%% file: table/cost.tex
\begin{table}[!ht]
\centering
\resizebox{0.52\textwidth}{!}{
\begin{tabular}{l>{\centering\arraybackslash}m{3cm}>{\centering\arraybackslash}m{2cm}>{\centering\arraybackslash}m{1cm}>{\centering\arraybackslash}m{2cm}>{\centering\arraybackslash}m{2cm}}
\Xhline{1pt}
\noalign{\smallskip}
\textbf{Method} & AUC@5/10/20 & \#Parameters(M) & GFLOPs & Latency(ms)\\
\cline{2-5}
\Xhline{1pt}
\noalign{\smallskip}
\textit{LoFTR} & 52.8 / 69.2 / 81.2 & 11.1 & 1767 & 281.6 \\
\textit{LoFTR-L} & 49.6 / 66.7 / 79.6 & 2.3 & 235 & 89.3 \\
\textbf{\textit{Ours-L}}  & 52.4 / 69.8 / 81.7 & 2.1 & 265 &  98.5 \\
\textit{QuadTree} & 54.6 / 70.5 / 82.2 & 13.2 & 1792 & 335.2  \\
\textit{ASpanFormer} & 55.3 / 71.5 / 83.1 & 15.5 & 1855 & 312.3  \\
\textbf{\textit{Ours}} & 57.3 / 72.8 / 84.0 & 12.8 & 1678 & 296.4 \\
\Xhline{1pt}
\end{tabular}
}
\caption{Performance-cost trade-off for different methods, we report auc on megadepth dataset. Ours-L denotes our light variants. Flops and latency for all methods are measured with image resize to 1200/1152 resolution. We use one V100 GPU for testing.}
\vspace{-1em}
\label{cost}
\end{table}

%% file: table/reg.tex
\begin{table}[!ht]
\centering
\resizebox{0.5\textwidth}{!}{
\begin{tabular}{l>{\centering\arraybackslash}m{1.2cm}>{\centering\arraybackslash}m{1.2cm}>{\centering\arraybackslash}m{1.2cm}}
\Xhline{1pt}
\noalign{\smallskip}
\multirow{2}{*}{\textbf{Design}}  & \multicolumn{3}{c}{\textbf{Pose Estimation AUC}} \\
\noalign{\smallskip}
\cline{2-4}
\noalign{\smallskip}
& @$5\degree$ & @$10\degree$ & @$20\degree$ \\
\noalign{\smallskip}
\Xhline{1pt}
\noalign{\smallskip}
~\textit{w. affine regularization} & 27.1 & 47.5 & 64.8 \\
~\textit{w/o affine regularization} & 26.5 & 46.9 & 63.9 \\
\Xhline{1pt}
\end{tabular}
}
\caption{Effect of affine regularization on ScanNet dataset~\cite{dai2017scannet}.}
\label{regularization}
\end{table}

%% file: cvpr.bbl
\begin{thebibliography}{10}\itemsep=-1pt

\bibitem{rootsift}
Relja Arandjelovi{\'c} and Andrew Zisserman.
\newblock Three things everyone should know to improve object retrieval.
\newblock In {\em CVPR}, 2012.

\bibitem{casmtr}
Chenjie Cao and Yanwei Fu.
\newblock Improving transformer-based image matching by cascaded capturing spatially informative keypoints.
\newblock 2023.

\bibitem{chen2022aspanformer}
Hongkai Chen, Zixin Luo, Lei Zhou, Yurun Tian, Mingmin Zhen, Tian Fang, David McKinnon, Yanghai Tsin, and Long Quan.
\newblock Aspanformer: Detector-free image matching with adaptive span transformer.
\newblock In {\em ECCV}, 2022.

\bibitem{choy2016universal}
Christopher~B Choy, JunYoung Gwak, Silvio Savarese, and Manmohan Chandraker.
\newblock Universal correspondence network.
\newblock {\em NeurIPS}, 2016.

\bibitem{dai2017scannet}
Angela Dai, Angel~X Chang, Manolis Savva, Maciej Halber, Thomas Funkhouser, and Matthias Nie{\ss}ner.
\newblock Scannet: Richly-annotated 3d reconstructions of indoor scenes.
\newblock In {\em CVPR}, 2017.

\bibitem{detone2018superpoint}
Daniel DeTone, Tomasz Malisiewicz, and Andrew Rabinovich.
\newblock Superpoint: Self-supervised interest point detection and description.
\newblock In {\em CVPRW}, 2018.

\bibitem{sp}
D. {DeTone}, T. {Malisiewicz}, and A. {Rabinovich}.
\newblock Superpoint: Self-supervised interest point detection and description.
\newblock In {\em CVPRW}, 2018.

\bibitem{dosovitskiy2020image}
Alexey Dosovitskiy, Lucas Beyer, Alexander Kolesnikov, Dirk Weissenborn, Xiaohua Zhai, Thomas Unterthiner, Mostafa Dehghani, Matthias Minderer, Georg Heigold, Sylvain Gelly, et~al.
\newblock An image is worth 16x16 words: Transformers for image recognition at scale.
\newblock In {\em ICLR}, 2020.

\bibitem{d2net}
Mihai Dusmanu, Ignacio Rocco, Tomas Pajdla, Marc Pollefeys, Josef Sivic, Akihiko Torii, and Torsten Sattler.
\newblock D2-net: A trainable cnn for joint description and detection of local features.
\newblock In {\em CVPR}, 2019.

\bibitem{roma}
Johan Edstedt, Qiyu Sun, Georg B{\"o}kman, M{\aa}rten Wadenb{\"a}ck, and Michael Felsberg.
\newblock Roma: Revisiting robust losses for dense feature matching.
\newblock {\em arXiv preprint arXiv:2305.15404}, 2023.

\bibitem{dkm}
Johan Edstedt, Mårten Wadenbäck, and Michael Felsberg.
\newblock Deep kernelized dense geometric matching.
\newblock {\em CVPR}, 2023.

\bibitem{topicfm}
Khang~Truong Giang, Soohwan Song, and Sungho Jo.
\newblock Topicfm: Robust and interpretable topic-assisted feature matching.
\newblock In {\em AAAI}, 2023.

\bibitem{resnet}
Kaiming He, Xiangyu Zhang, Shaoqing Ren, and Jian Sun.
\newblock Deep residual learning for image recognition.
\newblock In {\em CVPR}, 2016.

\bibitem{sfm2}
J. {Heinly}, J.~L. {Schönberger}, E. {Dunn}, and J. {Frahm}.
\newblock Reconstructing the world* in six days.
\newblock In {\em CVPR}, 2015.

\bibitem{dai17dcn}
Dai Jifeng, Qi Haozhi, Xiong Yuwen, Li Yi, Zhang Guodong, Hu Han, and Wei Yichen.
\newblock Deformable convolutional networks.
\newblock {\em ICCV}, 2017.

\bibitem{dai16rfcn}
Kaiming~He Jifeng~Dai, Yi~Li and Jian Sun.
\newblock {R-FCN}: Object detection via region-based fully convolutional networks.
\newblock 2016.

\bibitem{pats}
Ni Junjie, Li Yijin, Huang Zhaoyang, Li Hongsheng, Bao Hujun, Cui Zhaopeng, and Zhang Guofeng.
\newblock Pats: Patch area transportation with subdivision for local feature matching.
\newblock In {\em CVPR}, 2023.

\bibitem{li2020dual}
Xinghui Li, Kai Han, Shuda Li, and Victor Prisacariu.
\newblock Dual-resolution correspondence networks.
\newblock In {\em NeurIPS}, 2020.

\bibitem{li2018megadepth}
Zhengqi Li and Noah Snavely.
\newblock Megadepth: Learning single-view depth prediction from internet photos.
\newblock In {\em CVPR}, 2018.

\bibitem{mega}
Zhengqi Li and Noah Snavely.
\newblock Megadepth: Learning single-view depth prediction from internet photos.
\newblock In {\em CVPR}, 2018.

\bibitem{lightglue}
Philipp Lindenberger, Paul-Edouard Sarlin, and Marc Pollefeys.
\newblock Lightglue: Local feature matching at light speed.
\newblock 2023.

\bibitem{sift}
David~G Lowe.
\newblock Object recognition from local scale-invariant features.
\newblock In {\em ICCV}, 1999.

\bibitem{geodesc}
Zixin Luo, Tianwei Shen, Lei Zhou, Siyu Zhu, Runze Zhang, Yao Yao, Tian Fang, and Long Quan.
\newblock Geodesc: Learning local descriptors by integrating geometry constraints.
\newblock In {\em ECCV}, 2018.

\bibitem{luo2020aslfeat}
Zixin Luo, Lei Zhou, Xuyang Bai, Hongkai Chen, Jiahui Zhang, Yao Yao, Shiwei Li, Tian Fang, and Long Quan.
\newblock Aslfeat: Learning local features of accurate shape and localization.
\newblock In {\em CVPR}, 2020.

\bibitem{mishkin2018repeatability}
Dmytro Mishkin, Filip Radenovic, and Jiri Matas.
\newblock Repeatability is not enough: Learning affine regions via discriminability.
\newblock In {\em ECCV}, 2018.

\bibitem{mur2015orb}
Raul Mur-Artal, Jose Maria~Martinez Montiel, and Juan~D Tardos.
\newblock {ORB-SLAM}: a versatile and accurate monocular slam system.
\newblock {\em IEEE transactions on robotics}, 2015.

\bibitem{orbslam2}
Raul Mur-Artal and Juan Tardos.
\newblock {ORB-SLAM2}: an open-source slam system for monocular, stereo and rgb-d cameras.
\newblock {\em IEEE Transactions on Robotics}, 2016.

\bibitem{sparsencnet}
I. Rocco, R. Arandjelovi\'c, and J. Sivic.
\newblock Efficient neighbourhood consensus networks via submanifold sparse convolutions.
\newblock In {\em ECCV}, 2020.

\bibitem{ncnet}
I. Rocco, M. Cimpoi, R. Arandjelovi, A. Torii, T. Pajdla, and J. Sivic.
\newblock Neighbourhood consensus networks.
\newblock In {\em NeurIPS}, 2018.

\bibitem{sarlin2019coarse}
Paul-Edouard Sarlin, Cesar Cadena, Roland Siegwart, and Marcin Dymczyk.
\newblock From coarse to fine: Robust hierarchical localization at large scale.
\newblock In {\em CVPR}, 2019.

\bibitem{superglue}
Paul-Edouard Sarlin, Daniel DeTone, Tomasz Malisiewicz, and Andrew Rabinovich.
\newblock Superglue: Learning feature matching with graph neural networks.
\newblock In {\em CVPR}, 2020.

\bibitem{aachen}
Torsten Sattler, Will Maddern, Carl Toft, Akihiko Torii, Lars Hammarstrand, Erik Stenborg, Daniel Safari, Masatoshi Okutomi, Marc Pollefeys, Josef Sivic, Fredrik Kahl, and Tomas Pajdla.
\newblock Benchmarking 6dof outdoor visual localization in changing conditions.
\newblock In {\em CVPR}, 2018.

\bibitem{sattler2012imageBMCV}
Torsten Sattler, Tobias Weyand, Bastian Leibe, and Leif Kobbelt.
\newblock Image retrieval for image-based localization revisited.
\newblock In {\em BMVC}, 2012.

\bibitem{schonberger2016structure}
Johannes~L Schonberger and Jan-Michael Frahm.
\newblock Structure-from-motion revisited.
\newblock In {\em CVPR}, 2016.

\bibitem{sun2021loftr}
Jiaming Sun, Zehong Shen, Yuang Wang, Hujun Bao, and Xiaowei Zhou.
\newblock Loftr: Detector-free local feature matching with transformers.
\newblock In {\em CVPR}, 2021.

\bibitem{acnet}
Weiwei Sun, Wei Jiang, Eduard Trulls, Andrea Tagliasacchi, and Kwang~Moo Yi.
\newblock Acne: Attentive context normalization for robust permutation-equivariant learning.
\newblock In {\em CVPR}, 2020.

\bibitem{taira2018inloc}
Hajime Taira, Masatoshi Okutomi, Torsten Sattler, Mircea Cimpoi, Marc Pollefeys, Josef Sivic, Tomas Pajdla, and Akihiko Torii.
\newblock Inloc: Indoor visual localization with dense matching and view synthesis.
\newblock In {\em CVPR}, 2018.

\bibitem{tang2022quadtree}
Shitao Tang, Jiahui Zhang, Siyu Zhu, and Ping Tan.
\newblock Quadtree attention for vision transformers.
\newblock In {\em ICLR}, 2021.

\bibitem{truong2021pdc}
Prune Truong, Martin Danelljan, Luc~Van Gool, and Radu Timofte.
\newblock Learning accurate dense correspondences and when to trust them.
\newblock In {\em CVPR}, 2021.

\bibitem{pdcnet+}
Prune Truong, Martin Danelljan, Radu Timofte, and Luc Van~Gool.
\newblock {PDC-Net+}: Enhanced probabilistic dense correspondence network.
\newblock {\em Preprint}, 2021.

\bibitem{transformer}
Ashish Vaswani, Noam Shazeer, Niki Parmar, Jakob Uszkoreit, Llion Jones, Aidan~N Gomez, \L~ukasz Kaiser, and Illia Polosukhin.
\newblock Attention is all you need.
\newblock In {\em NeurIPS}. 2017.

\bibitem{wang2022matchformer}
Qing Wang, Jiaming Zhang, Kailun Yang, Kunyu Peng, and Rainer Stiefelhagen.
\newblock Matchformer: Interleaving attention in transformers for feature matching.
\newblock 2022.

\bibitem{caps}
Qianqian Wang, Xiaowei Zhou, Bharath Hariharan, and Noah Snavely.
\newblock Learning feature descriptors using camera pose supervision.
\newblock In {\em ECCV}, 2020.

\bibitem{eloftr}
Yifan Wang, Xingyi He, Sida Peng, Dongli Tan, and Xiaowei Zhou.
\newblock {Efficient LoFTR}: Semi-dense local feature matching with sparse-like speed.
\newblock In {\em CVPR}, 2024.

\bibitem{xia2022vision}
Zhuofan Xia, Xuran Pan, Shiji Song, Li~Erran Li, and Gao Huang.
\newblock Vision transformer with deformable attention.
\newblock In {\em CVPR}, 2022.

\bibitem{lift}
Kwang~Moo Yi, Eduard Trulls, Vincent Lepetit, and Pascal Fua.
\newblock Lift: Learned invariant feature transform.
\newblock In {\em ECCV}, 2016.

\bibitem{yi2016lift}
Kwang~Moo Yi, Eduard Trulls, Vincent Lepetit, and Pascal Fua.
\newblock Lift: Learned invariant feature transform.
\newblock In {\em ECCV}, 2016.

\bibitem{yi2016learning}
Kwang~Moo Yi, Yannick Verdie, Pascal Fua, and Vincent Lepetit.
\newblock Learning to assign orientations to feature points.
\newblock In {\em CVPR}, 2016.

\bibitem{yu2023adaptive}
Jiahuan Yu, Jiahao Chang, Jianfeng He, Tianzhu Zhang, and Feng Wu.
\newblock Adaptive spot-guided transformer for consistent local feature matching.
\newblock {\em CVPR}, 2023.

\bibitem{astr}
Jiahuan Yu, Jiahao Chang, Jianfeng He, Tianzhu Zhang, Jiyang Yu, and Wu Feng.
\newblock {ASTR}: Adaptive spot-guided transformer for consistent local feature matching.
\newblock {\em CVPR}, 2023.

\bibitem{oanet}
Jiahui Zhang, Dawei Sun, Zixin Luo, Anbang Yao, Lei Zhou, Tianwei Shen, Yurong Chen, Long Quan, and Hongen Liao.
\newblock Learning two-view correspondences and geometry using order-aware network.
\newblock In {\em ICCV}, 2019.

\bibitem{zhang2021reference}
Zichao Zhang, Torsten Sattler, and Davide Scaramuzza.
\newblock Reference pose generation for long-term visual localization via learned features and view synthesis.
\newblock {\em IJCV}, 2021.

\bibitem{aliked}
Xiaoming Zhao, Xingming Wu, Weihai Chen, Peter~CY Chen, Qingsong Xu, and Zhengguo Li.
\newblock Aliked: A lighter keypoint and descriptor extraction network via deformable transformation.
\newblock {\em IEEE Transactions on Instrumentation and Measurement}, 2023.

\bibitem{alike}
Xiaoming Zhao, Xingming Wu, Jinyu Miao, Weihai Chen, Peter~CY Chen, and Zhengguo Li.
\newblock Alike: Accurate and lightweight keypoint detection and descriptor extraction.
\newblock {\em IEEE Transactions on Multimedia}, 2022.

\bibitem{kfnet}
Lei Zhou, Zixin Luo, Tianwei Shen, Jiahui Zhang, Mingmin Zhen, Yao Yao, Tian Fang, and Long Quan.
\newblock Kfnet: Learning temporal camera relocalization using kalman filtering.
\newblock In {\em CVPR}, 2020.

\bibitem{zhu2020deformable}
Xizhou Zhu, Weijie Su, Lewei Lu, Bin Li, Xiaogang Wang, and Jifeng Dai.
\newblock Deformable detr: Deformable transformers for end-to-end object detection.
\newblock {\em ICLR}, 2020.

\end{thebibliography}
